\documentclass[runningheads]{llncs}
\usepackage{graphicx}
\usepackage{subcaption}
\usepackage{algorithm}
\usepackage{algorithmic}
\usepackage{amsmath}
\begin{document}

\title{GroupMixNorm Layer for Learning Fair Models}
%
\author{Anubha Pandey\orcidID{0000-0002-4695-0947} \and
Aditi Rai\orcidID{0009-0009-9298-7861} \and 
Maneet Singh \and
Deepak Bhatt\orcidID{0000-0003-3694-1315} \and
Tanmoy Bhowmik
}
\authorrunning{A. Pandey et al.}
\institute{AI Garage, Mastercard, India \\
\email{\{anubha.pandey, aditi.rai, maneet.singh, deepak.bhatt\}@mastercard.com, tantanmoy@gmail.com}}

%
\maketitle              
\begin{abstract}

Recent research has identified discriminatory behavior of automated prediction algorithms towards groups identified on specific protected attributes (e.g., gender, ethnicity, age group, etc.). When deployed in real-world scenarios, such techniques may demonstrate biased predictions resulting in unfair outcomes. Recent literature has witnessed algorithms for mitigating such biased behavior mostly by adding convex surrogates of fairness metrics such as demographic parity or equalized odds in the loss function, which are often not easy to estimate. This research proposes a novel in-processing based \textit{GroupMixNorm} layer for mitigating bias from deep learning models. The GroupMixNorm layer probabilistically mixes group-level feature statistics of samples across different groups based on the protected attribute. The proposed method improves upon several fairness metrics with minimal impact on overall accuracy. Analysis on benchmark tabular and image datasets demonstrates the efficacy of the proposed method in achieving state-of-the-art performance. Further, the experimental analysis also suggests the robustness of the GroupMixNorm layer against new protected attributes during inference and its utility in eliminating bias from a pre-trained network. 




\end{abstract}

\section{Introduction}


Most AI algorithms process large quantities of data to identify patterns useful for accurate predictions. Such pipelines are mostly automated in nature without any human intervention, along with large data processing, high efficiency, and high accuracy. Despite the benefits of automated processing, current AI systems are marred with the challenge of biased predictions resulting in unfavourable outcomes. One of the most infamous examples of such behavior is that of an AI-based recruitment tool\footnote{\url{https://tinyurl.com/5apv7xeu}}, which disfavoured applications from women because it was trained on resumes from the mostly male workforce. 
In order to rectify such biases and support advancement in society, we need models that generate fair results without any discrimination towards certain individuals or groups. To this effect, this research proposes a novel \textit{GroupMixNorm} layer for learning an unbiased model for ensuring fair outcomes across different groups.

In the literature, research has focused on achieving fairness by introducing techniques at the pre-processing stage (transforming the input before feeding to the classification model) or the post-processing stage (transforming the output produced by the classification model). It is our hypothesis that these methods may not result in optimal accuracy, since they treat the classifier as a black box and focus on removing bias from the input representations or the output predictions only. Different from the above, \textit{in-processing} techniques focus on learning bias-invariant models by incorporating additional constraints during training, thus resulting in more effective models \cite{fairMLBook}. Existing in-processing techniques mostly aim to solve a constraint optimization problem to ensure fairness \cite{inprocessing1,inprocessing2,Adversary} by introducing a penalty term in the loss function corresponding to the convex surrogates of the fairness objective like \textit{demographic parity} or \textit{equalized odds}. However, as observed in literature, it is challenging to formulate surrogates for different fairness constraints that is a reasonable estimate of the original \cite{fnnc}. 


\begin{figure}[t]
  \centering
  \includegraphics[scale=0.1]{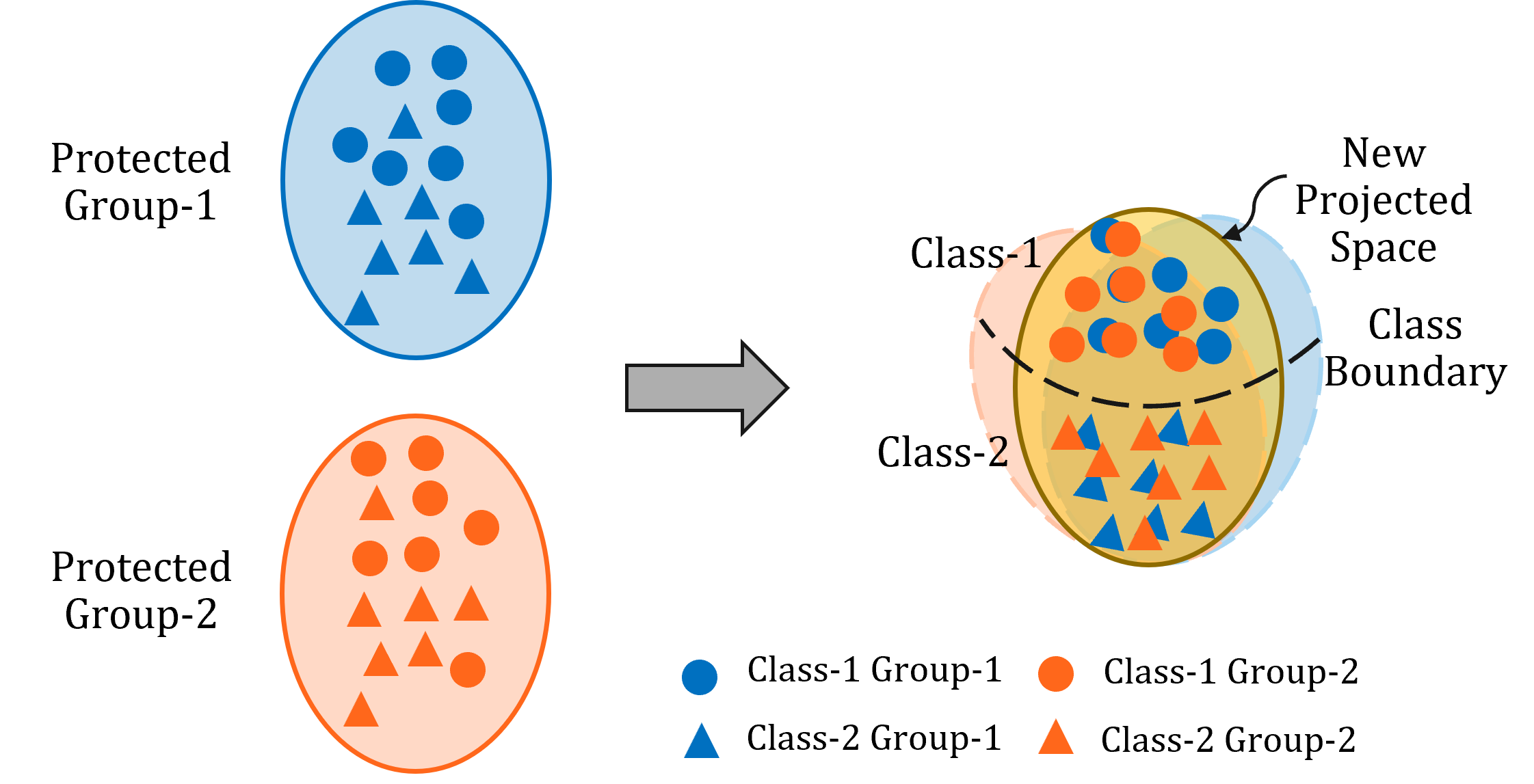}
  \caption{The proposed GroupMixNorm layer projects the features of different classes and protected attributes onto a space which minimizes the distinction between the protected attributes, thus promoting a fairer classification model.}
  \label{fig:intro_fig}
    \vspace{-15pt}
\end{figure}

In this research, we formulate the problem of bias mitigation as distribution alignment of several groups of the protected attribute (Fig. \ref{fig:intro_fig}). The proposed \textit{GroupMixNorm} layer is applied at the in-processing stage which promotes the model to learn unbiased features for classification. The formulation is motivated by the observation that Deep Learning based algorithms tend to explore the difference in the distribution among the groups of the protected attributes (e.g., male and female with similar features like age and education may have different salaries, thus resulting in different distributions) to lift the overall performance. The GroupMixNorm layer mixes the group-level feature statistics and transforms all the features in a training batch based on the interpolated group statistics. This enables the classifier to learn features invariant to the protected attribute. Further, transforming the data towards the interpolated groups regularizes the classifier and improves the generalizability at inference. Key highlights of this research are as follows:
\begin{itemize}
    \item This research proposes a novel \textit{GroupMixNorm} layer for learning fairer classification models. The proposed layer is applied at the architectural level and is an in-processing technique that focuses on distribution alignment of different groups during model training. 
    
    \item GroupMixNorm operates at the feature level, thus making it flexible to be placed across various layers of a neural network-based model and fits well into the mini-batch gradient-based training. Experimental analysis suggests that with limited data, GroupMixNorm can be applied to mitigate the existing bias in classifiers as well, thus avoiding the need for re-training from scratch.
    
    \item The GroupMixNorm layer produces fairer results when evaluated for new groups at test time as well. We believe that the GroupMixNorm layer makes the model robust against distribution changes across sensitive groups, thus being able to generalize well for unseen groups at test time.
    
    \item The efficacy of the proposed approach has been demonstrated on different datasets (structured and unstructured), where it achieves improved performance while achieving multiple fairness constraints such as demographic parity, equal opportunity, and equalized odds simultaneously. For example, on the UCI Adult Income dataset \cite{UCI}, GroupMixNorm achieves an average precision of 0.77, while maintaining different fairness metrics below 0.03. 
    
\end{itemize}


\section{Related Work}

Group fairness can be ensured in a machine learning system via \textit{pre-processing}, \textit{in-processing}, and \textit{post-processing}. 
Pre-processing and post-processing methods consider the classifier as a black-box model, and try to mitigate bias from the input features or the classifier's prediction. 
\begin{figure*}[t]
  \centering
\includegraphics[scale=0.1]{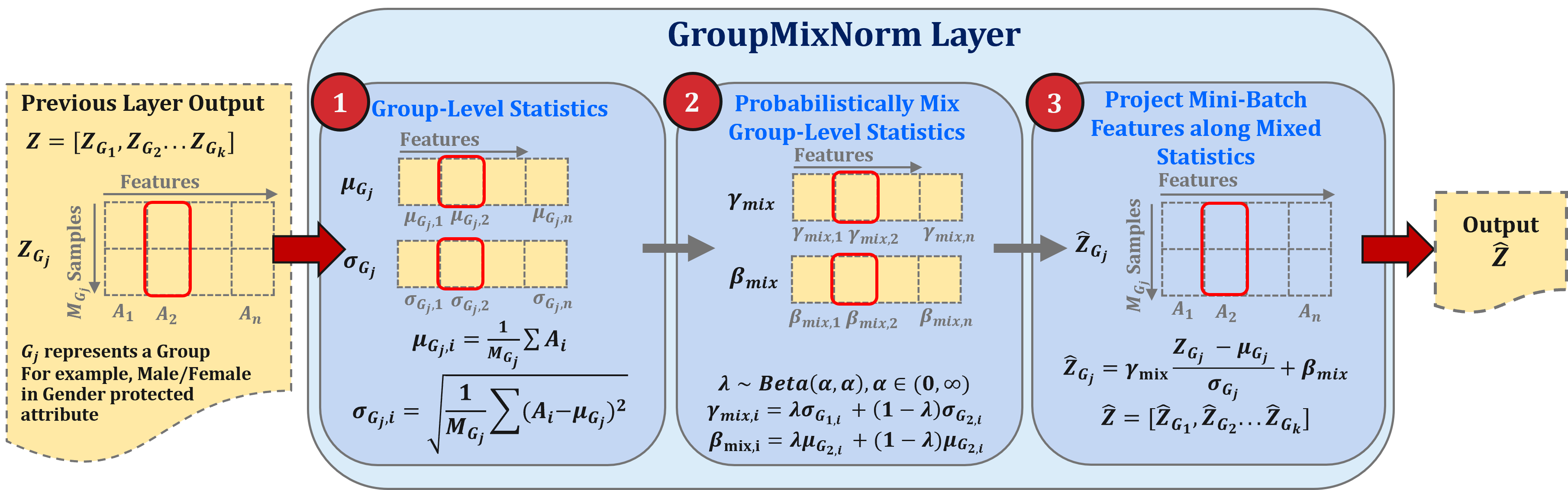}
\caption{The GroupMixNorm layer takes as input the previous layer's output (${\mathbf{Z}}$) along with each sample's protected attribute. Group-level statistics are computed, followed by the probabilistic mixing and projection of the mini-batch features along the mixed statistics to obtain new features ($\mathbf{\hat{Z}}$).}
\label{fig:GroupMix}
\vspace{-15pt}
\end{figure*}
On the other hand, \textit{in-processing} based bias mitigation techniques solve the constraint optimization problem for different fairness objectives. To ensure independence between the predictions and sensitive attributes, Woodworth et al. \cite{inprocessing1} regularize the covariance between them. Zafar et al. \cite{inprocessing2} minimize the disparity between the sensitive groups by regularizing the decision boundary of the classifier. 
Game theory based approaches \cite{inprocessing_gen1,inprocessing_gen2} provide analytical solutions and theoretical guarantees for generalizability in fair classifier but are limited by the scalability factor. Recent techniques \cite{adv_learn1,Adversary} introduce an adversary network additional to the predictor network that predicts the sensitive label based on the classifier's output, while other algorithms 
\cite{inv_risk2,inv_risk1,model_explain1,model_explain2} learn unbiased representations through invariant risk minimization and attention-based feature learning. Research has also focused on eliminating superficial correlations and paying more attention on task related causal features \cite{causalFairness2,causalFairness1}. Recently, Cheng et al. \cite{Contrastive} utilize contrastive learning to minimize the correlation between sentence representations and biasing words, while mixup \cite{verma2019manifold,mixup} techniques have proved to be effective in bias mitigation. For example, Chuang et al. \cite{fairmixup} utilize mixup as a data augmentation strategy to improve the generalizability of the model while optimizing the fairness constraints, and Du et al. \cite{rnf} utilize mixup for feature neutralization to remove the correlation between the sensitive information and class labels from the encoder feature.

Instead of focusing on optimizing surrogates of the fairness metrics, this research proposes a novel GroupMixNorm layer which operates at the \textit{architectural} level of the classifier. GroupMixNorm focuses on learning unbiased representations which results in satisfying several fairness constraints across groups.




\section{Proposed GroupMixNorm Layer}


As discussed before, recent research has observed that deep learning models often tend to learn group-specific characteristics, making it easier to obtain a higher performance on the underlying classification task. As an ancillary effect, the learned group-specific features often also result in discriminative behavior towards specific groups based on the protected attribute. For example, a recruitment tool may learn features based on the gender of the applicant, resulting in unintended discrimination towards applicants from the under-represented group. In order to address the above limitation, the GroupMixNorm layer focuses on eliminating the difference between the group statistics during training.

As part of the GroupMixNorm layer, we normalize each group of a protected attribute in a batch separately to collect group specific statistics (i.e. for the gender attribute, normalize all male samples and female samples in a batch separately) and further take a probabilistic convex combination between the group-level statistics and apply across all the samples in a batch. This process ensures that any protected group related diversity is removed from the internal representation of a neural network and doesn't allow the network to explore this information to lift the overall performance. The introduction of additional inductive bias in the network structure enforces it to learn invariant features pertaining to the protected attributes while training the network.

The GroupMixNorm layer is implemented as a plug-and-play module. It can be inserted between the fully connected layers of a neural network-based classifier during training (Algorithm \ref{alg:algorithm}). Let $X, Y$, and $S$ be the input features, class labels, and protected attribute labels in a training batch, respectively. As illustrated in Fig. \ref{fig:GroupMix}, let $Z$ be an $n$ dimensional representation obtained from the previous layer and $A_i$ represent the feature along dimension $i$. We identify the groups $G_j$ in a batch based on the protected attribute labels $S$, and calculate their respective mean ($\mu_{G_j,i}$) and variance ($\sigma_{G_j,i}$) along each dimension (step-1 of Fig. \ref{fig:GroupMix}). Next we calculate the weighted average of mean $\gamma_{mix,i}$ and variance $\beta_{mix,i}$ along each dimension (Eq. \ref{eqn:var_mix_i}), followed by concatenation to create a single vector (Eq. \ref{eqn:mean_mix_i}). As we mix statistics of two groups at a time, the mixing coefficient $\lambda$ is sampled from a symmetric Beta distribution $Beta(\alpha, \alpha)$, for $\alpha \in (0, \infty)$. The hyper-parameters $\alpha$ controls the strength of interpolation.

Finally, we normalize all the samples by applying the calculated $\gamma_{mix}$ and $\beta_{mix}$ to each sample as shown in Eq. \ref{eqn:group_mix}. For the ease of notation, we have considered two groups i.e. binary protected attributes. However, the proposed solution can easily be applied to non-binary protected attributes as well. 

\begin{equation}\label{eqn:var_mix_i}
\gamma_{mix,i} = \lambda \sigma_{G_1,i} + (1-\lambda) \sigma_{G_2,i}; \ \beta_{mix,i} = \lambda \mu_{G_1,i} + (1-\lambda) \mu_{G_2,i}; \end{equation}
\begin{equation}\label{eqn:mean_mix_i}
\gamma_{mix} = [\gamma_{mix,1},... \gamma_{mix,n}]; \ \beta_{mix} = [\beta_{mix,1},... \beta_{mix,n}]
\end{equation}
\begin{equation}\label{eqn:group_mix}
\hat{Z}_{G_j} = \gamma_{mix} \frac{(Z_{G_j}-\mu_{G_j})}{\sigma_{G_j}}+ \beta_{mix}
\end{equation}

The updated features $\hat{Z} = [\hat{Z}_{G_1}, \hat{Z}_{G_2}]$ are then provided as input to the following layer of the neural network for further processing. The process of mixing group level statistics in a GroupMixNorm layer occurs in the feature space and has no learnable parameters. The GroupMixNorm layer is easy to implement and fits perfectly into mini-batch training. 
Further, it is turned off during inference, thus eliminating the need for protected attributes during inference. The training procedure of GroupMixNorm layer is shown in Algorithm \ref{alg:algorithm}.

\begin{algorithm}[tb]
\caption{GroupMixNorm Layer}
\label{alg:algorithm}
\textbf{Input}:
$Z$: Learned representation of the input batch obtained from the previous layer\\
$\alpha, \beta$: Hyper-parameters for the Beta distribution (default: 0.1) \\
\textbf{Output}: $\hat{Z}$: Transformed samples after the GroupMixNorm layer

\begin{algorithmic}[1] 
\IF{not in training mode}
\STATE \textbf{return} $Z$ \
\ENDIF
\STATE Compute $\mu_{G_{j}}$ and $\sigma_{G_{j}}$ for a group $G_j$ in a protected attribute \
\STATE Sample mixing coefficient $\lambda \sim Beta(\alpha, \alpha)$ 
\STATE Compute $\gamma_{mix}$ and $\beta_{mix}$ as shown in Eq. \ref{eqn:mean_mix_i}
\STATE Normalize and transform all samples in a batch to compute $\hat{Z}_{G_j}$ as shown in Eq. \ref{eqn:group_mix}
\STATE $\hat{Z} = \{\hat{Z}_{G_j}\}_{j=1}^{K}$, where $K$ is the number of groups identified in a protected attribute
\STATE \textbf{return} $\hat{Z}$
\end{algorithmic}
\end{algorithm}
\vspace{-15pt}

\section{Datasets and Experimental Details}

The GroupMixNorm layer has been evaluated on two datasets with different fairness evaluation metrics and compared with state-of-the-art techniques. Details regarding the dataset protocols are as follows:  
\begin{itemize}
    \item \textbf{UCI Adult Dataset \cite{UCI}} contains 50,000 samples with 14 attributes to describe each data point (individual) (e.g., gender, education level, age, etc.) from the 1994 US Census. The classification task is to predict the income of an individual. It's a binary classification task, where class 1 represents salary $\geq$ 50K and class 0 represents salary $<$ 50K. We select gender as the protected attribute for the fairness evaluation. The dataset is imbalanced such that only $24\%$ of the samples belong to class 1, with only $15.13\%$ female samples.
    \item \textbf{CelebA Dataset \cite{celeba}} contains 200,000 celebrity faces with 40 binary attributes associated with each image. Following the literature \cite{fairmixup,rnf}, we select gender as the protected attribute and wavy hair attribute for the binary classification task. The dataset has $18.36\%$ male samples as compared to female samples in the positive class.
\end{itemize}

\subsection{Fairness Evaluation Metrics}
\label{section:fair_eval_metrics}
The most widely used fairness metrics  \cite{fairnessDefinition} are: Demographic Parity, Equal Opportunity, and Equalized Odds \cite{EOP}. The metrics are elaborated in detail below, where $Y$ ($\widehat{Y}$) is actual (predicted) class label and $S$ is protected attribute:

\begin{itemize}

\item \textbf{Demographic Parity Difference (DP)} suggests that the probability of favourable outcomes should be same for all the subgroups:
\begin{equation}
\label{eqn:dpd}
DPD = | P[\widehat{Y} = 1|S = 1] - P[\widehat{Y}  = 1 | S\neq 1]|
\end{equation}

\item \textbf{Equality of Opportunity Difference (EOP)} emphasises that there should be equal opportunities for all the subgroups having positive outcomes to have positive prediction i.e. true positive rates for all the groups should be same:
\begin{equation}
\label{eqn:eop}
EOP = | P[\widehat{Y} = 1|S = 1,Y = 1] - 
P[\widehat{Y}  = 1 | S\neq 1,Y = 1]| 
\end{equation}
\item \textbf{Equalized Odds Difference (EOD)} focuses on equalizing false positive rates along with the same true positive rates for all the subgroups:
\begin{equation}
\label{eqn:aod}
\begin{aligned}
EOD = |P[\widehat{Y} = 1|S = 1,Y = 1] - P[\widehat{Y}  = 1 | S\neq 1,Y = 1]| + \\
|P[\widehat{Y} = 1|S = 1,Y = 0] - P[\widehat{Y}  = 1 | S\neq 1,Y = 0])|
\end{aligned}
\end{equation}
For a fair algorithm, DP, EO and EOD values must be closer to 0.
\end{itemize}

\subsection{Implementation Details}

The GroupMixNorm layer has been implemented in the PyTorch framework on Ubuntu 16.04.7 OS with the Nvidia GeForce GTX 1080Ti GPU. For a fair comparison with existing literature, we have followed the same dataset pre-processing and protocols as the fair mixup approach \cite{fairmixup}. For the Adult dataset, we use four fully connected layers with hidden dimension 50. Each layer except the last output layer is followed by SiLU activation and the proposed GroupMixNorm layer. The model is trained for 10 epochs with a 1000 batch size. For each epoch, the dataset is randomly split into 60-20-20 split of train, val, and test set, respectively. We select the best-performing model on the validation set across 10 independent runs and report the mean Average Precision and fairness metrics defined above. For the CelebA dataset, we use the ResNet-18 \cite{resnet} model for feature extraction followed by two fully connected layers for the classification task. We apply SiLU activation and GroupMixNorm layer between the two FC layers. We use the original split of the dataset and train the model for 100 epochs with 128 batch size. Both the models are trained with the Adam optimizer with learning rate $1e-4$. In all experiments, mixing coefficient $\lambda$ (Eqs. \ref{eqn:var_mix_i} and \ref{eqn:mean_mix_i}) is randomly sampled from $Beta(\alpha, \alpha)$. The value of $\alpha$ is empirically set to 0.1.



\section{Results and Analysis}

\begin{figure*}[!t]
\centering
\subcaptionbox{Demographic Parity}{\includegraphics[width=0.33\textwidth]{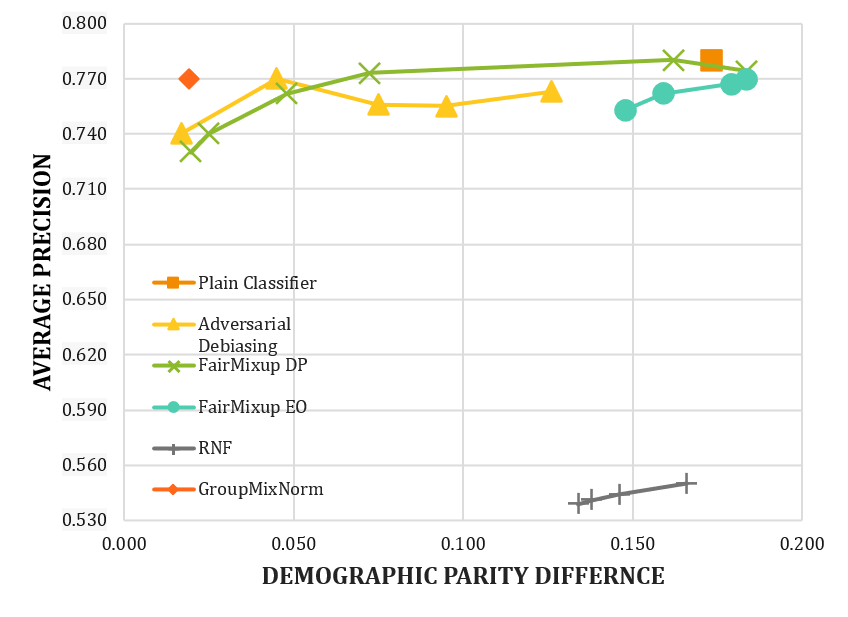}}%
\hfill 
\subcaptionbox{Equal Opportunity}{\includegraphics[width=0.33\textwidth]{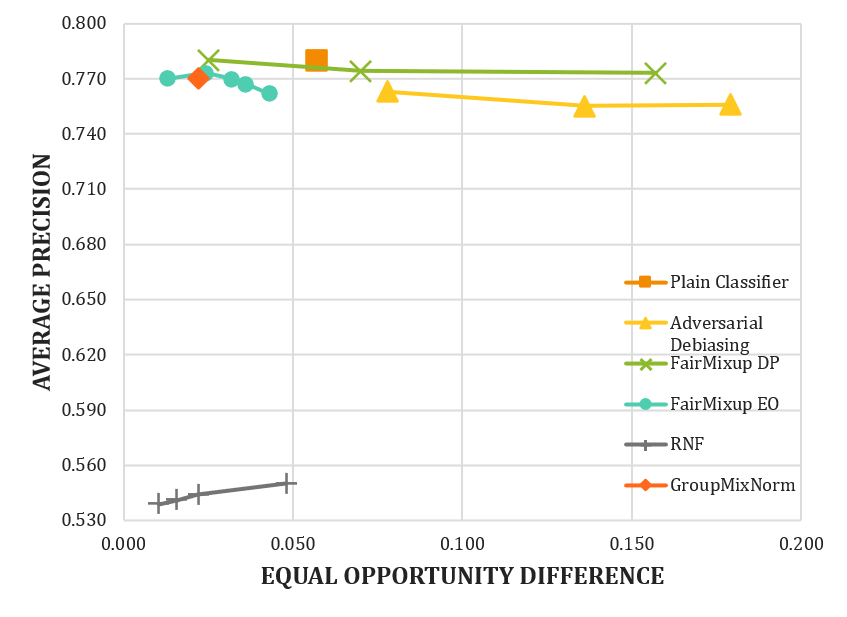}}%
\hfill 
\subcaptionbox{Equalized Odds}{\includegraphics[width=0.33\textwidth]{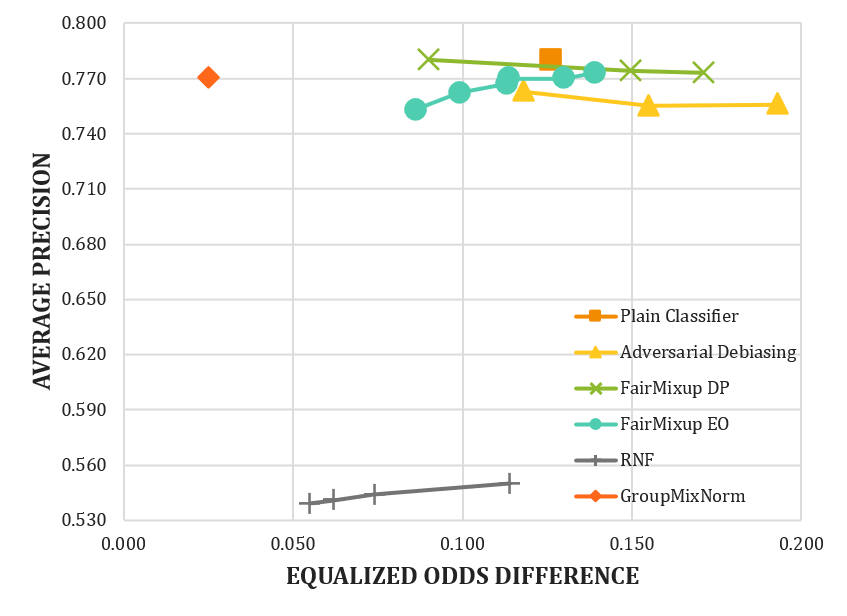}}%
\caption{\textbf{Fairness-AP trade-off curves} on the \textbf{Adult dataset}, where GroupMixNorm demonstrates improved performance. Results are obtained by varying the trade-off parameter as suggested in their respective publications: Adversarial Debiasing: [0.01 $\sim$ 1.0], Fair Mixup DP: [0.1 $\sim$ 0.7], Fair Mixup EO: [0.5 $\sim$ 5.0], and RNF: [0.05, 0.015, 0.025, 0.035]. For a fair algorithm, it is desirable to have the AP closer to 1, and the fairness metrics (DP, EO, EOD) closer to 0.}
\label{fig:Adult_tradeoff}
\vspace{-15pt}
\end{figure*}

\begin{figure*}[!t]
\centering
\subcaptionbox{Demographic Parity}{\includegraphics[width=0.33\textwidth]{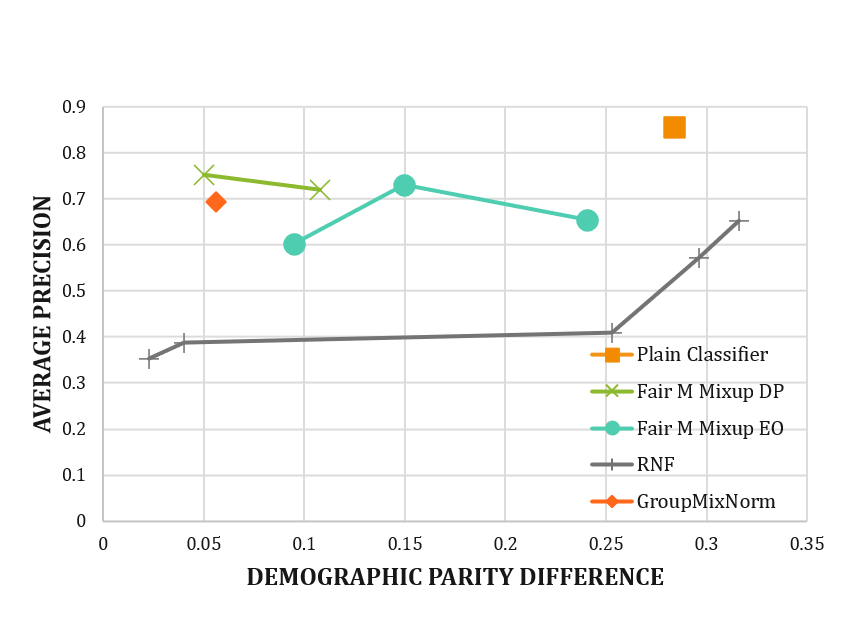}}%
\hfill 
\subcaptionbox{Equal Opportunity}{\includegraphics[width=0.33\textwidth]{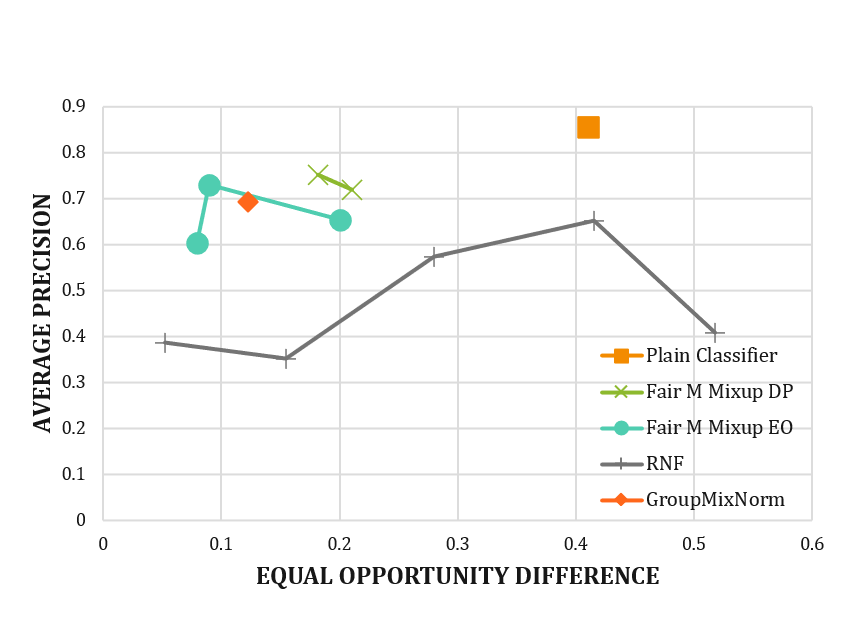}}%
\hfill 
\subcaptionbox{Equalized Odds}{\includegraphics[width=0.33\textwidth]{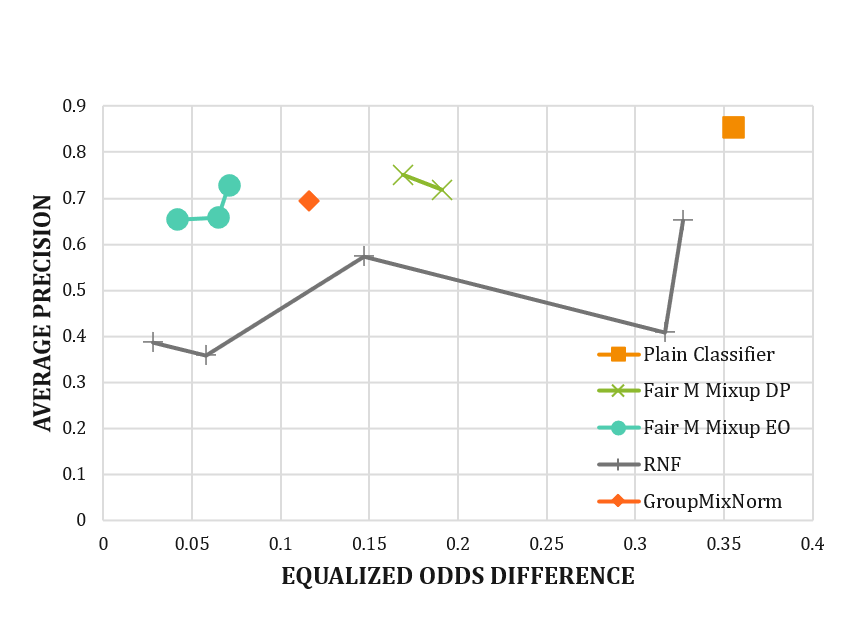}}%
\caption{\textbf{Fairness-AP trade-off curves} of  GroupMixNorm layer and other comparative algorithms on the \textbf{CelebA dataset}, where the proposed approach achieves state-of-the-art performance. Results are obtained by varying the trade-off parameter as suggested in their respective publications: Fair M Mixup DP: [25, 50], Fair M Mixup EO: [1, 10, 50], RNF: [0.1, 0.5, 1, 5, 10].} 
\label{fig:celeb_tradeoff}
\vspace{-15pt}
\end{figure*}

\begin{figure*}[!t]
\centering
\begin{subfigure}{0.49\textwidth}
    \centering
    \includegraphics[trim={1cm, 1cm, 1cm, 2cm}, clip, width=1\linewidth]{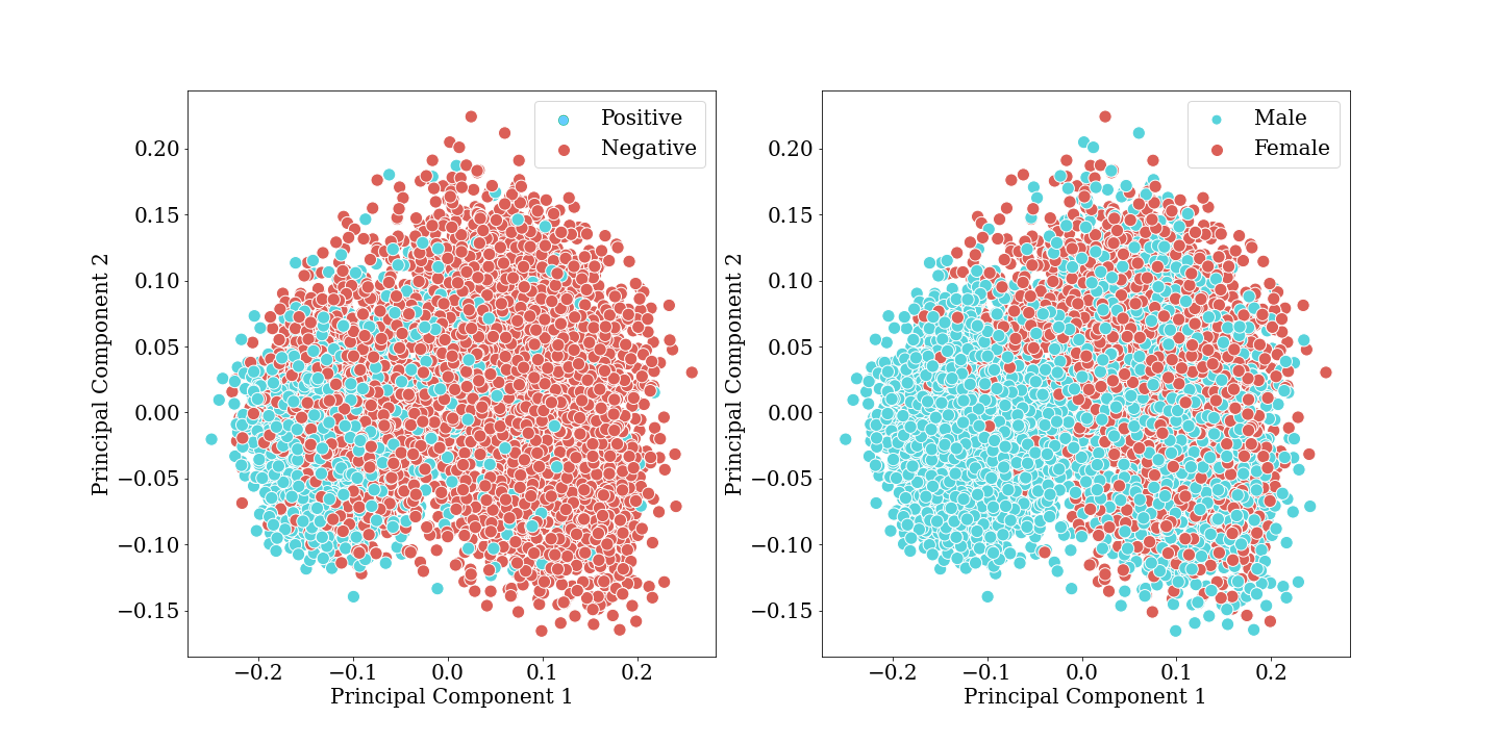}  
    \caption{Without GroupMixNorm Layer}
    \label{fig:plaincls_feat}
\end{subfigure}
\begin{subfigure}{0.49\textwidth}
    \centering
    \includegraphics[trim={1cm, 1cm, 1cm, 2cm}, clip, width=1\linewidth]{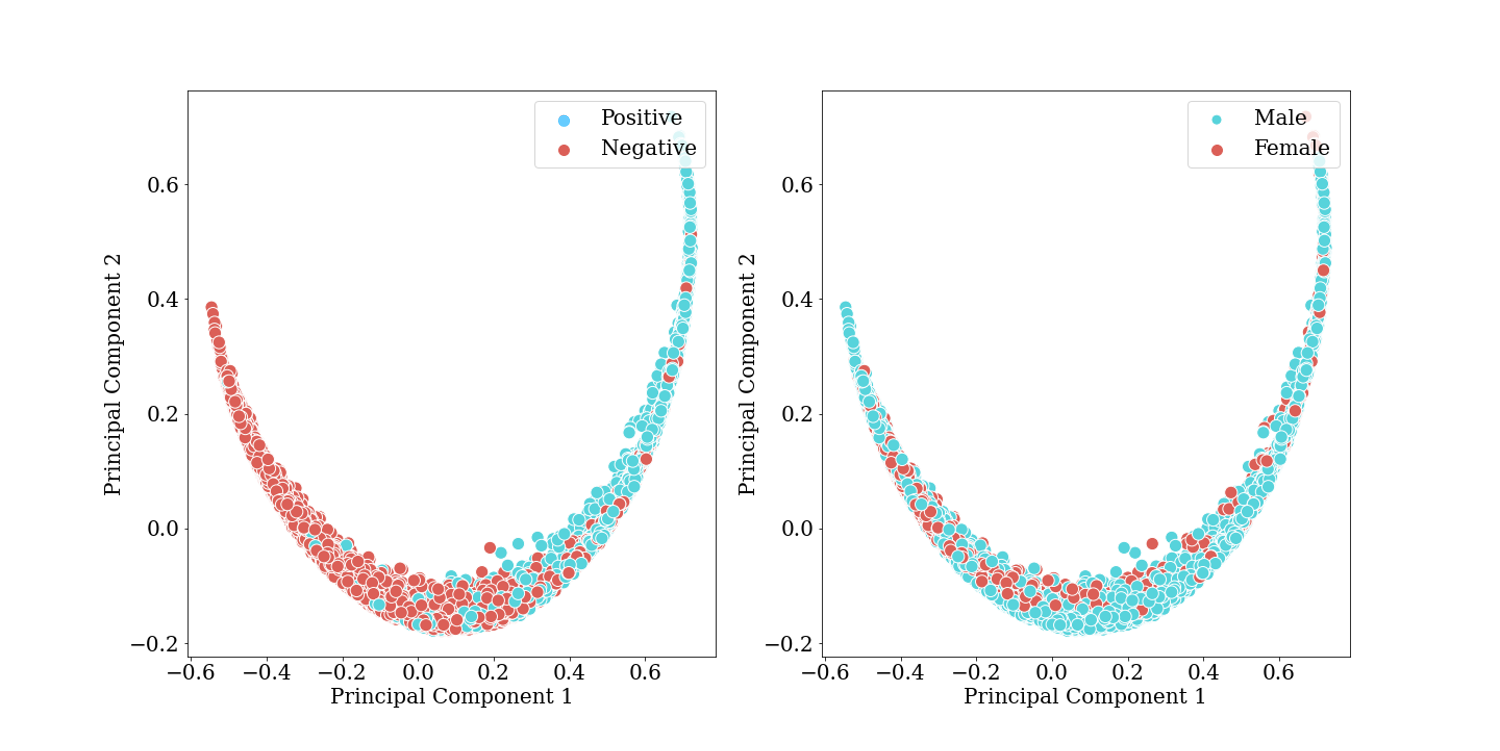}  
    \caption{With GroupMixNorm Layer}
    \label{fig:mixgroupnorm_feat}
\end{subfigure}
\caption{PCA visualizations of the features for the MLP classifiers trained without and with GroupMixNorm. The left plots show the class distribution, and the right plots show the gender distribution (protected attribute). The model trained with GroupMixNorm demonstrates minimal distinction on the gender attribute.}
\label{fig:pca_features}
\vspace{-15pt}
\end{figure*}

Figures \ref{fig:Adult_tradeoff}--\ref{fig:Adult_zeroShot} and Table \ref{tab:cosine_sim} present the results and analysis of the GroupMixNorm layer and comparison with the state-of-the-art in-processing bias mitigation techniques. Detailed analysis is given in the following subsections: 

\subsection{Comparison with State-of-the-art Algorithms}
Since the GroupMixNorm layer focuses on mitigating bias during the training process, comparison has been performed with algorithms that optimize fairness constraints during training: (i) Adversarial Debiasing \cite{Adversary}, (ii) Fair Mixup: Fairness via Interpolation (Fair Mixup) \cite{fairmixup}, (iii) Fairness via Representation Neutralization (RNF) \cite{rnf}, and (iv) plain classifier. Fair Mixup uses two separate regularizing terms for optimizing the fairness metrics of Demographic Parity (DP) and Equal Opportunity (EO), and thus can solve for either DP or EO at a time. In this paper, we refer to these two variants of Fair Mixup as \textit{Fair Mixup DP} and \textit{Fair Mixup EO}. 
To calculate the DP, Chuang \textit{et al.} \cite{fairmixup} have computed the difference between the predicted probability across the protected groups. Similarly, for EO, Chuang \textit{et al.} \cite{fairmixup} compute class-wise difference between the predicted probability across protected groups. As part of this research, we have used the actual definitions of EO and DP for computing the fairness metrics (Eqs. \ref{eqn:dpd}--\ref{eqn:eop}). Parallely, the Representation Neutralization (RNF) technique \cite{rnf} has shown the bias mitigation performance via two variants: (i) in model-1, proxy labels are generated for the protected attribute, while (ii) in model-2, ground-truth protected attribute labels are used. As part of this research, we have compared our results with their second variant (model-2), referred to as \textit{RNF}. 

For a fair comparison, we evaluate all the models under the same setting. 
Techniques such as Adversarial Debiasing, Fair Mixup, and RNF introduce a regularization term in the loss function to improve fairness via a hyper-parameter $\alpha$ that controls the trade-off between the average precision (AP) and fairness metrics (DP, EO, and EOD). We have reported the results on varying values of $\alpha$ as suggested in their respective papers. In our case, the GroupMixNorm layer is proposed towards architecture design and not the loss function, thus there is no such trade-off. Performance analysis on different datasets is as follows:

\noindent \textbf{(a) Comparison on the UCI Adult Income Dataset:} Fig. \ref{fig:Adult_tradeoff} shows the performance comparison on the UCI Adult dataset, where the GroupMixNorm layer produces fairer results as compared to other techniques across all fairness metrics (DP, EO, EOD) with minimal impact on average precision. Since Fair Mixup solves separate constraint optimizations to achieve lower DP and EO, it minimizes either DP or EO at a time. In terms of the fairness metrics, RNF produces fair results, however the average precision is relatively lower, thus making it unsuitable for the classification task. 

\noindent  \textbf{(b) Comparison on the CelebA Dataset (Fig. \ref{fig:celeb_tradeoff}):} Consistent with the published manuscript \cite{fairmixup}, for Fair Mixup, comparison has been performed with the combination of manifold mixup \cite{verma2019manifold} (Fair M Mixup DP and Fair M Mixup EO). Similar to the previous experiments, it is observed that either Fair M Mixup DP or Fair M Mixup EO achieves optimal performance at a time. Further, the RNF model produces fair results across fairness metrics, however achieves lower average precision. GroupMixNorm achieves comparable performance to the best performing model across all the metrics, while maintaining a high average precision, thus suggesting high utility for real-world applications.

\subsection{Learned Representation Analysis}

Experiments have been performed for (a) feature visualization and (b) auxiliary prediction task for understanding feature quality. The key findings are as follows:
 
\noindent \textbf{(a) Feature Visualization:} Fig. \ref{fig:pca_features} presents the 2D projections obtained by using the sigmoid kernel Principal Component Analysis (PCA). Fig. \ref{fig:plaincls_feat} presents the features learned by a biased MLP classifier (trained without GroupMixNorm layer), where the features appear both class and gender (protected attribute) discriminative. There is an overlap of male samples with the positive class samples, both lying majorly in the lower left side of the distribution. On the other hand, Fig. \ref{fig:mixgroupnorm_feat} shows features learned with the GroupMixNorm layer appear to be class discriminative, while not being gender discriminative. Further, both male and female samples are evenly distributed, thus preventing the model to get biased against a particular sensitive group.

\begin{table}[t]
\caption{Cosine similarity between the learned weight parameters of $C_{sens}$ and $C_{cls}$ linear classifiers (the former is trained for predicting the sensitive attribute, while the later is trained for class label prediction). A lower score represents less biased models since lesser similarity is observed between the weight parameters.}
\centering
\begin{tabular}{|l|c|}
\hline
\textbf{Method}  & \textbf{Cosine Similarity} \\
\hline
\hline
Plain Classifier       & 0.205     \\
\hline
RNF       & 0.075-0.2      \\
\hline
GroupMixNorm    & 0.06     \\
\hline
\end{tabular}
\label{tab:cosine_sim}
\vspace{-10pt}
\end{table}

\noindent \textbf{(b) Auxiliary Prediction Task:} Similar to Du \textit{et al.} \cite{rnf}, we use an auxiliary prediction task to analyze the quality of the learned features. The objective is to analyze how well the model can reduce the correlation between the class labels and the sensitive attributes. To this effect, we train two linear classifiers $C_{sens}$ and $C_{cls}$ that take the representation vector as input and predicts class labels and sensitive attributes, respectively. Next, we compare the learned weight matrix of $C_{sens}$ and $C_{cls}$ using cosine similarity. A higher similarity would signify similar weights and thus higher correlation between the two tasks. Table \ref{tab:cosine_sim} shows that our model has the least cosine similarity indicating that the classifier focuses more on task relevant information than sensitive information. It is important to note that the cosine similarity for the RNF model varies from 0.2 to 0.075, based on the fairness-accuracy trade-off parameter, while the GroupMixNorm layer based model achieves a cosine similarity of 0.06 only.

\subsection{Generalizability to New Protected Groups}
\begin{figure*}[!t]
\centering
\begin{subfigure}{0.49\textwidth}
    \centering
    \includegraphics[trim={0cm, 0cm, 0cm, 0cm}, clip, width=0.80\linewidth]{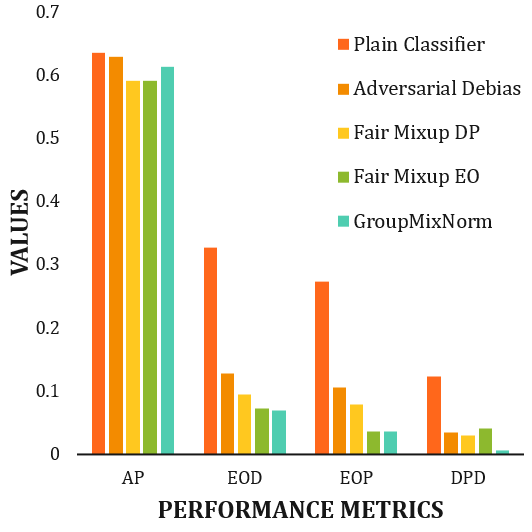}  
    \caption{New Protected Attributes}
    \label{fig:newProtAttribute}
\end{subfigure}
\begin{subfigure}{0.49\textwidth}
    \centering
    \includegraphics[trim={0cm, 0cm, 0cm, 0cm}, clip, width=0.80\linewidth]{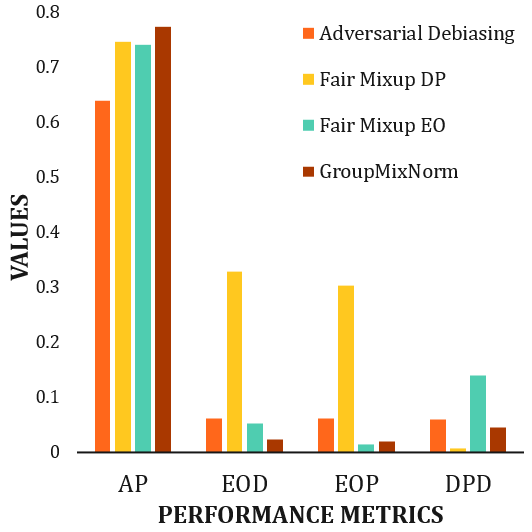}  
    \caption{De-biasing Pre-trained Classifier}
\label{fig:Adult_intra-procesing}
\end{subfigure}
\vspace{-5pt}
\caption{Average precision and fairness metrics obtained by different techniques (a) when evaluated on new protected attributes and (b) for de-biasing a pre-trained classifier with limited training data. Experiments have been performed on the Adult Income dataset with race and gender as the protected attribute, respectively. GroupMixNorm presents improved performance across metrics. }
\label{fig:Adult_zeroShot}
\vspace{-15pt}
\end{figure*}

With time, as the data evolves, new sensitive groups often get introduced. For example, gender attribute values may change from binary to non-binary. A robust classification model must remain unbiased even with the introduction of additional sensitive groups during inference. In order to simulate this setup, the proposed solution was evaluated for new groups at test time without any re-training. Experiments were performed on the Adult Income dataset where data pertaining to two races (White and Black) was used for training, while the data from White, Black, and Others (Asian-Pac-Islander, Amer-Indian-Eskimo, Other) racial groups were used for testing. Fig. \ref{fig:Adult_zeroShot}a presents the performance of the GroupMixNorm layer along with other comparative techniques, where GroupMixNorm is able to generalize well to unseen groups during inference by obtaining lower fairness metrics and a higher average precision.

\subsection{Debias Pre-trained Model with Limited Data}

Experiments have also been performed to analyze the effectiveness of the proposed GroupMixNorm layer to mitigate bias from a pre-trained biased classifier. We train an MLP classifier on the training partition of the Adult Income dataset, without the GroupMixNorm layer, and later fine-tune the model after plugging the proposed layer on the validation set. The validation set consists of only $20\%$ samples of the entire dataset. For other techniques, we fine-tune the pre-trained biased classifier on the validation set with the respective methods. We evaluate the model for fairness on the Adult dataset with gender as the protected attribute. Fig. \ref{fig:Adult_intra-procesing} presents the results obtained by the proposed GroupMixNorm layer as well as other comparative techniques. It can be observed that the proposed solution produces fairer results as compared to other algorithms across the different fairness metrics, while achieving the highest average precision. The experiment suggests that even with a small training set, the proposed GroupMixNorm can aid in eliminating bias from a pre-trained network. 




\section{Conclusion and Future Work}

Learning bias-invariant models are the need of the hour for the research community. While existing research has focused on proposing novel solutions for learning unbiased classifiers, most of the techniques incorporate an additional term in the loss function for modeling the model fairness. We believe that it is often difficult to extrapolate the learnings of such an optimization function to the test set, especially under the challenging scenario of new protected attributes during evaluation. To this effect, this research proposes a novel \textit{GroupMixNorm} layer, which promotes learning fairer models at the architectural level. GroupMixNorm is a distribution alignment strategy operating across the different protected groups, enabling attribute-invariant feature learning. 
Across multiple experiments, GroupMixNorm demonstrates improved fairness metrics while maintaining higher average precision levels, as compared to the state-of-the-art algorithms. Further analysis suggests high model generalizability to new protected attributes during evaluation, possibly due to the transformation of samples to interpolated groups resulting in model regularization during training. As an extension of this research, future research directions include studying the impact of GroupMixNorm on different convolution layers and extending the scope to evaluation on NLP datasets and tasks. 

\bibliographystyle{splncs04}
\bibliography{ref}

\end{document}